%% file: rap-2025-inverse-dynamics-solver.tex
\documentclass[journal]{IEEEtran} 

\usepackage{amsmath,amsfonts}
\usepackage{algorithmic}
\usepackage{algorithm}
\usepackage{array}
\usepackage[caption=false,font=normalsize,labelfont=sf,textfont=sf]{subfig}
\usepackage{textcomp}
\usepackage{stfloats}
\usepackage{url}
\usepackage{verbatim}
\usepackage{graphicx}
\usepackage{cite}
\usepackage[hidelinks]{hyperref}
\usepackage{bm}
\usepackage[dvipsnames]{xcolor}
\usepackage{tikz}
\input{scripts/tikz-config}

\usepackage{xspace}
\newcommand*{\eg}{e.g.\@\xspace}
\newcommand*{\ie}{i.e.\@\xspace}
\newcommand*{\wrt}{w.r.t.\@\xspace}
\newcommand*{\ms}{M.Sc.\@\xspace}  
\newcommand*{\phd}{Ph.D.\@\xspace}  
\newcommand*{\dr}{Dr.\@\xspace}  
\newcommand*{\prof}{Prof.\@\xspace}  

\newcommand*{\gh}{\url{https://github.com/unisa-acg/inverse-dynamics-solver/tree/rap}}
\newcommand*{\co}{\url{https://doi.org/10.24433/CO.2265930.v1}}
\newcommand*{\pr}{\url{https://github.com/ros/rosdistro/pull/44979}}
\newcommand*{\gbp}{\url{https://github.com/ros2-gbp/ros2-gbp-github-org/issues/732}}

\begin{document}

\title{
A ROS2-based software library for inverse dynamics computation
}

\author{
Vincenzo Petrone,~\IEEEmembership{Student Member,~IEEE,}
Enrico Ferrentino,~\IEEEmembership{Member,~IEEE,} and
Pasquale Chiacchio
\thanks{
Authors are with the Department of Information Engineering, Electrical Engineering and Applied Mathematics (DIEM), University of Salerno, 84084 Fisciano, Italy (email: \{vipetrone, eferrentino, pchiacchio\}@unisa.it).
}
}

\markboth{
IEEE Robotics and Automation Practice
}%
{
Petrone \MakeLowercase{\textit{et al.}}: A ROS2-based software library for inverse dynamics computation
}

\IEEEpubid{0000--0000/00\$00.00~\copyright~2021 IEEE}

\maketitle

\begin{abstract}
Inverse dynamics computation is a critical component in robot control, planning and simulation, enabling the calculation of joint torques required to achieve a desired motion.
This paper presents a ROS2-based software library designed to solve the inverse dynamics problem for robotic systems.
The library is built around an abstract class with three concrete implementations: one for simulated robots and two for real UR10 and Franka robots.
This contribution aims to provide a flexible, extensible, robot-agnostic solution to inverse dynamics, suitable for both simulation and real-world scenarios involving planning and control applications.
The related software is available at \gh.
\end{abstract}


\section{Introduction}

\IEEEPARstart{M}{odeling} robotic systems' dynamics is a fundamental aspect in the design and control of robots \cite{carpentier_pinocchio_2019}.
In this context, the computation of \emph{inverse dynamics} plays a critical role.
Inverse dynamics involves determining the torques required at each joint of a robot to produce a specified motion, given its joint positions, velocities, and accelerations \cite{featherstone_dynamics_2016}.

This calculation is essential for a wide range of applications, from optimal motion planning \cite{ferrentino_dynamic_2024} to advanced control \cite{antonelli_systematic_1999} and human-robot collaboration \cite{gaz_model-based_2018}.
Accurate and efficient computation of inverse dynamics is particularly critical in applications such as high-performance control systems \cite{adam_safety_2024}, trajectory optimization \cite{petrone_time-optimal_2022}, and real-time simulations \cite{puck_performance_2021}.

In the ROS2 framework \cite{macenski_robot_2022} -- a de-facto standard in developing robotics software \cite{dehnavi_compros_2021,fernando_facile_2024,richard_omnilrs_2024} -- a common library that exposes interfaces to such paramount components does not currently exist.
Releasing and documenting a software to fill this gap is the focus of this paper.

\subsection{Background}\label{sec:background}

The inverse dynamics problem can be framed as the determination of the dynamics components acting on a robotic system, according to the following model \cite{featherstone_dynamics_2016}:
\begin{equation}\label{eq:eom}
    \bm H(\bm q) \ddot{\bm q} + \bm C(\bm q, \dot{\bm q})\dot{\bm q} + \bm f(\dot{\bm q}) + \bm g(\bm q) = \bm\tau,
\end{equation}
where $\bm H$ is the inertia term, $\bm C$ accounts for Coriolis and centrifugal effects, $\bm f$ is the friction vector, $\bm g$ describes the torques due to gravity, and $\bm\tau$ are the torques applied to the system, while $\bm q$, $\dot{\bm q}$ and $\ddot{\bm q}$ indicate joint positions, velocities and accelerations, respectively.

The inverse dynamics model requires a detailed representation of these components to compute the torques $\bm\tau$ necessary to achieve a specified joint motion, described by the kinematic variables $(\bm q, \dot{\bm q}, \ddot{\bm q})$.
Usually, the dynamic terms in \eqref{eq:eom} are described through equations of motion (EOM), typically devised according to Newton-Euler or Lagrange formulations, as a function of joint motion variables and parametrized \wrt DH kinematic parameters and dynamic parameters \cite{hollerbach_model_2016}.

\subsection{Contribution}\IEEEpubidadjcol
\label{sec:contribution}

\input{scripts/dynamics}

The primary contribution of this paper is a \emph{ROS2-based software library} designed to compute inverse dynamics for robotic systems, as in \eqref{eq:eom}.
Our module addresses the need for a flexible, extensible, and robot-agnostic solution to the inverse dynamics problem.
The library is built around an abstract class called \emph{Inverse Dynamics Solver} (IDS) that defines a generic interface for inverse dynamics computation, ensuring its broad applicability across various robot platforms.

As a further contribution, we provide three concrete implementations of this abstract class, as illustrated in Fig.~\ref{fig:inverse-dynamics}:
\begin{enumerate}
    \item \textbf{A generic implementation for simulated robots}, based on KDL, suitable for use in physics-based simulators such as Gazebo or others in the ROS ecosystem.
    This implementation leverages the kinematic and dynamic parameters of the simulated robot to calculate the relevant dynamics components.
    \item \textbf{Two implementations for real robots}, tailored for specific robotic platforms, namely the the popular UR10 and Franka manipulators.
    These implementations account for real-world dynamics, including nonlinear friction models,  making them suitable for applications involving optimal planning, accurate control, and hardware-in-the-loop testing.
\end{enumerate}

The primary goal of this library is to provide a robust and reusable tool for computing inverse dynamics in both simulated and real-world scenarios, with particular emphasis on ease of integration within ROS2-based systems.
By adopting a modular approach with neat abstractions, our library aims to simplify the integration of inverse dynamics solvers into a wide range of contexts involving robotic applications.

\section{Inverse Dynamics Solver}\label{sec:inverse-dynamics-solver}

\input{figures/ids-sequence-diagram}

This section presents the functionality and architecture of the library, so as to provide a structural and behavioral description to the community.
The readers can refer to our public repository for further details\footnote{\gh}.

\subsection{Functionality}\label{sec:functionalities}

The IDS library exposes functions to allow the user to extract $\bm H(\bm q)$, $\bm C(\bm q,\dot{\bm q})\dot{\bm q}$, $\bm f(\dot{\bm q})$, $\bm g(\bm q)$ or $\bm\tau$.
By default, $\bm\tau$ is computed as in \eqref{eq:eom} by setting $\bm f(\dot{\bm q}) \equiv 0$.
This choice is driven by the fact that, differently from the other components, it usually does not depend on linear dynamic parameters (typically estimated via identification techniques \cite{petrone_dynamic_2024, gaz_dynamic_2019}), and its formulation might follow various linear or nonlinear models \cite{gaz_model-based_2018}. Furthermore, some modern robot controllers compensate friction internally \cite{fci, Chawda_2017}, relieving the user from accounting for it in the design of the control software \cite{Le_Tien_2008}.

The library follows the \texttt{pluginlib} paradigm\footnote{\url{https://github.com/ros/pluginlib/tree/ros2}}, \ie, the software does not strictly depend on the concrete implementation, which can be chosen via configuration, \eg via launch files.
This critical design choice favors the robot-agnosticism required by off-the-shelf ROS libraries.

At instance time, the robot description is required, which can be easily retrieved by parsing the URDF file with \texttt{xacro}\footnote{\url{http://wiki.ros.org/urdf}}.
Furthermore, the solver can be configured by choosing the \texttt{root} and the \texttt{tip} of the chain to solve the dynamics for, as well as the \texttt{gravity} vector, which influences the $\bm g(\bm q)$ component.
The functioning of the IDS library and its communication mechanisms with external entities in the ROS2 framework are schematized in Fig.~\ref{fig:sequence-diagram} with a sequence diagram.

\subsection{Architecture}

\input{figures/ids-class-diagram}

As shown in Fig.~\ref{fig:class-diagram}, the software architecture is composed by an abstract class, named \texttt{InverseDynamicsSolver}, providing the functionalities mentioned in Section~\ref{sec:functionalities}: this class represents a generic solver, and the particular implementation of the methods returning dynamic components is a responsibility of the concrete classes.

\subsubsection{Abstract class}\label{sec:interface}

The interface exposes the following public methods, all inherited by the concrete implementations:
\begin{itemize}
    \item \texttt{initialize}: initializes the solver with the robot description and the \texttt{root}, \texttt{tip} and \texttt{gravity} parameters through the node parameters interface;
    \item \texttt{getInertiaMatrix}: returns $\bm H(\bm q)$;
    \item \texttt{getCoriolisVector}: returns $\bm C(\bm q, \dot{\bm q})\dot{\bm q}$;
    \item \texttt{getFrictionVector}: returns $\bm f(\dot{\bm q})$;
    \item \texttt{getGravityVector}: returns $\bm g(\bm q)$;
    \item \texttt{getDynamicComponents}: returns $\left( \bm H, \bm C\dot{\bm q}, \bm g \right)$
    \item \texttt{getTorques}: returns $\bm\tau = \bm H(\bm q)\ddot{\bm q} + \bm C(\bm q, \dot{\bm q})\dot{\bm q} + \bm g(\bm q)$.
\end{itemize}
As mentioned in Section~\ref{sec:contribution}, this class is derived by three solvers we provide.
The KDL-based one is tailored for simulated robots, and discussed in Section~\ref{sec:kdl}.
Two solvers for real robots, specifically UR10 and Franka, are presented in Section~\ref{sec:solvers}. 
It is worth highlighting that users in the ROS2 community are allowed to inherit the base class with their own implementations, possibly including the estimated models of other manipulators.

\subsubsection{KDL-based solver}\label{sec:kdl}

By exploiting the KDL library\footnote{\url{https://github.com/orocos/orocos_kinematics_dynamics}}, the \texttt{InverseDynamicsSolverKDL} concrete class computes the dynamic components, acting as a bridge between KDL's and our IDS' interfaces.
Specializing the \texttt{initialize} method, it parses the robot description to build the kinematic chain via KDL.
This class is useful when a complete kinematic and dynamic description is available, \eg in the case both the DH and dynamic parameters are provided by the manufacturer, to be specified in the URDF file.
However, it disregards the friction term $\bm f$ since, as mentioned in Section~\ref{sec:background}, it does not depend on the inertial parameters.
Therefore, the users are invited to adopt this library in simulation only, where joint friction can be disregarded.

\subsubsection{Real UR10 and Franka solvers}\label{sec:solvers}

We provide a solver for the UR10 industrial manipulator: the \texttt{InverseDynamicsSolverUR10} class returns the torques for this robot, according to the model estimated by \cite{petrone_dynamic_2024}, including the nonlinear friction vector, and a regressive form for other terms.
In \cite{petrone_dynamic_2024}, the robot is identified at current level, hence this class exposes methods to retrieve the joint currents, while the torque-based methods exploit the \texttt{getDriveGainsMatrix} function to convert currents to torques (see Fig.~\ref{fig:class-diagram}).
Additionally, we share a solver for the Franka research robot: the \texttt{InverseDynamicsSolverFrankaInria} class implements the model estimated by \cite{gaz_dynamic_2019} which, differently from the \texttt{libfranka} \cite{haddadin_franka_2022} implementation provided by the manufacturer, also provides a formulation for the nonlinear friction.
Given the presence of $\bm f$, both these solvers use \eqref{eq:eom} to compute $\bm\tau$, hence overloading the default \texttt{getTorques} method defined in the parent class, differently from the KDL solver discussed in Section~\ref{sec:kdl}.

\subsection{Implementation details}

The IDS defines the input/output types of the implemented methods using \texttt{Eigen}\footnote{\url{https://eigen.tuxfamily.org/dox}}, a popular C++ linear algebra  library, widely used in ROS2.
This way, users can easily manipulate the computed matrices for model-based applications, such as writing control schemes or evaluating planning algorithms \cite{petrone_time-optimal_2022}.

\section{Results}

This section reports the results obtained by employing the IDS.
To demonstrate its robot-agnosticism, we adopt the same interface to retrieve the dynamic components of 4 robots, namely simulated and real UR10 and Franka manipulators.
It is worth remarking that providing accurate dynamic models falls outside the scope of this article, as the correctness of the computed torques relies on the underlying estimations \cite{petrone_dynamic_2024, gaz_dynamic_2019}.
Thus, the goal of this section is demonstrating the application of our IDS in different scenarios across various robotic platforms.
Indeed, the results in terms of computed torques are \emph{all} obtained with the \emph{same} demo, exclusively re-configured with different plugins.
The readers are invited to replicate the results with the reproducible CodeOcean capsule\footnote{\co}.

\subsection{Simulations}

\input{figures/kdl-ur10/kdl-ur10}

\input{figures/kdl-franka/kdl-franka}

We exercise the KDL-based solver described in Section~\ref{sec:kdl} in a Gazebo simulation: we command a trajectory to a simulated UR10 robot via a \texttt{ros2\_control} joint trajectory controller, and compute the torques corresponding to each joint state with the IDS' \texttt{getTorques} method, which computes all the dynamics components in \eqref{eq:eom}, as explained in Section~\ref{sec:interface}.
Fig.~\ref{fig:kdl-ur10} shows the joint torques returned by the IDS along the trajectory.
Similarly, we command a trajectory to the Franka robot, and evaluate the same solver to retrieve the results displayed in Fig.~\ref{fig:kdl-franka}.
Notably, in both cases measured and computed torque signals showcase a perfect match, as Gazebo itself uses KDL to simulate the system dynamics.
Consequently, the \texttt{InverseDynamicsSolverKDL} can be easily plugged in a robot-agnostic torque-based controller, to reliably compensate the dynamics of any simulated robot.

\subsection{Experiments}

\subsubsection{UR10}\label{sec:ur10}

\input{figures/ur10/ur10}

Commanding an excitation trajectory to the real UR10 robot through the ROS2 driver\footnote{\url{https://github.com/UniversalRobots/Universal_Robots_ROS2_Driver}}, we measure the actual joint configurations and currents via a utility provided by the manufacturer\footnote{\url{https://github.com/UniversalRobots/RTDE_Python_Client_Library}}, and obtain the corresponding torques with the \texttt{getDriveGainsMatrix} method (see Section~\ref{sec:solvers}).
Comparing them with the torques computed through our \texttt{InverseDynamicsSolverUR10} library, Fig.~\ref{fig:ur10} shows that they closely match the ones retrieved from actual currents, demonstrating the correctness of the model in \cite{petrone_dynamic_2024} and of our software.

\subsubsection{Franka}

\input{figures/franka/franka}

Similarly to what described in Section~\ref{sec:ur10} for the UR10 manipulator, we command an excitation trajectory to the real Franka robot, and evaluate the \texttt{InverseDynamicsSolverFrankaInria} library on each measured joint configuration.
Fig.~\ref{fig:franka} shows the measured torques, obtained via \texttt{libfranka} \cite{haddadin_franka_2022}, against the ones computed with our IDS library.
We remark that, as demonstrated in \cite{gaz_dynamic_2019}, the model we chose to implement in our software provides an explicit estimation of the friction contribution, differently from the manufacturer interface.
The advantage of this choice is that, although this manipulator compensates for friction internally in control applications \cite{fci}, having an explicit model is crucial for, \eg, offline planning purposes \cite{petrone_time-optimal_2022, ferrentino_dynamic_2024}.

\section{Conclusions}

This paper presented a ROS2-based robot-agnostic library to solve the inverse dynamics problem.
Given joint positions, velocities and accelerations, the developed software can be used to compute the torques acting on the manipulator's joints, along with the dynamic terms influencing the EOM.

We released and documented a codebase including an interface, exposing methods to retrieve the aforementioned dynamic components, and three concrete implementations, implementing the estimated models of two popular real manipulators, and adopting the KDL library for simulated robots.
With the abstract class we shared, other users in the ROS2 community can specialize the library with new concrete classes, implementing the solvers for different robotic platforms. 

The off-the-shelf library can be used in planning algorithms, control schemes, and simulations, enriching the community with a useful tool to adopt in various applications requiring the estimation or compensation of the robot dynamics.
The software is validated in both simulated and real scenarios, to demonstrate its functioning in devising the torques of the manipulators we exercised our library on.

In the future, we intend to extend our codebase by including the information about the robot's payload, which might influence the dynamics \cite{gaz_payload_2017}, or implementing an additional concrete solver based on the \texttt{Pinocchio} library \cite{carpentier_pinocchio_2019}.
Furthermore, we plan to include the solver in model-based controllers, \eg inverse dynamics or gravity compensation controllers, to be released as separate packages.

{
\appendix[Description of Supplementary Materials]

This work is supported by the following supplementary materials, detailed in the next sections:
\begin{itemize}
\item a video, showing our validation setup and animating the obtained results: \url{https://youtu.be/p5ITcvoozGc};
\item a CodeOcean capsule, to reproduce our results: \co;
\item a GitHub repository, containing the ROS2 packages: \gh.
\end{itemize}

\subsection{Video}
The attached video shows the execution of 4 excitation trajectories on simulated and real robots, namely Franka and UR10 manipulators (see Fig.~\ref{fig:robots}).
The plots included in Fig.~\ref{fig:kdl-ur10}--\ref{fig:franka} are animated, to show the measured and computed torque signals along the robots' movements.

\input{figures/robots/robots}

\subsection{Code}

\subsubsection{CodeOcean capsule}

By accessing our capsule on CodeOcean, users can execute the IDS software via formal unit tests and demos.
On the one hand, tests are developed to validate the main functionalities of the library (see Sect.~\ref{sec:inverse-dynamics-solver}), \eg parameter-based initialization and correctness of the computed torques.
On the other hand, demos replicate the results displayed in Fig~\ref{fig:kdl-ur10}--\ref{fig:franka}.
It is worth stressing that the CodeOcean capsule does not merely plot the \textit{same} data again, but actually builds and runs the concrete solvers from scratch, producing \textit{new} data (\ie, torque predictions) that, eventually, are plotted to reproduce the aforementioned graphs.

\subsubsection{GitHub repository}

To strengthen the impact of our work, we opened a Pull Request (PR) from our GitHub repository, so that the developed packages will be subject to code review from other members of the ROS2 community, and eventually released as part of ROS2 distributions.
The PR towards \texttt{rosdistro}, to index our package under the \texttt{humble} distribution, is reachable at \pr.
Furthermore, we opened an issue to ask to release our packages under ROS2 Git Built Package (GBP), reachable at \gbp.
Upon acceptance, our packages will be distributed via \texttt{rosdep}.

}

\bibliographystyle{IEEEtran}
\bibliography{rap-2025-inverse-dynamics-solver}

\end{document}

%% file: scripts/tikz-config.tex
\usetikzlibrary{positioning, shadows}

\newlength{\blockWidth}
\newlength{\blockHeight}
\newlength{\arrowLength}
\newlength{\cornerRadius}
\newcommand*{\labelFontSize}{\small}

\definecolor{cadmiumgreen}{rgb}{0.0, 0.42, 0.24}
\definecolor{ao(english)}{rgb}{0.0, 0.5, 0.0}

\tikzset{
    simple text/.style={
        draw=none,
        font=\footnotesize,
    },
    square/.style={
        draw,
        ultra thick,
        minimum width=\blockHeight,
        minimum height=\blockHeight,
        rounded corners=\cornerRadius,
        fill=white,
        drop shadow,
    },
	block/.style={
		draw,
		ultra thick,
		minimum width=\blockWidth,
		minimum height=\blockHeight,
		rounded corners=\cornerRadius,
        text width=0.75\blockWidth,
        align=center,
        fill=BrickRed,
        text=white,
        drop shadow,
	},
	arrow/.style={
		draw,
		very thick,
		-stealth,
	},
    input arrow/.style={
        arrow,
        MidnightBlue,
    },
    output arrow/.style={
        arrow,
        ao(english),
    }
}

\definecolor{lgblue}{RGB}{0, 113.985, 188.955}
\definecolor{lgred}{RGB}{216.750, 82.875, 24.990}
\definecolor{lgyellow}{RGB}{236.895, 176.970, 31.875}

\newcommand{\defineleg}[2]{
    \newcommand{#1}{\raisebox{2pt}{\tikz{\draw[#2, line width=0.9pt](0,0)--(6mm,0);}}}
}

\defineleg{\legblack}{black, dashed}
\defineleg{\legred}{red}

%% file: scripts/dynamics.tex
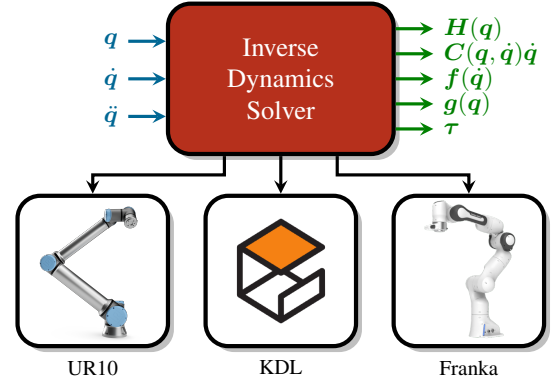
\begin{figure}
\centering

\begin{tikzpicture}

\node[block] (ids) at (0,0) {Inverse Dynamics Solver};

\draw[input arrow] ([xshift=-\arrowLength,yshift=+0.25\blockHeight]ids.west) -- node[pos=0, left, font=\labelFontSize]{$\bm q$} ++(\arrowLength,0);
\draw[input arrow] ([xshift=-\arrowLength]ids.west) -- node[pos=0, left, font=\labelFontSize]{$\dot{\bm q}$} ++(\arrowLength,0);
\draw[input arrow] ([xshift=-\arrowLength,yshift=-0.25\blockHeight]ids.west) -- node[pos=0, left, font=\labelFontSize]{$\ddot{\bm q}$} ++(\arrowLength,0);

\draw[output arrow] ([yshift=\blockHeight/3]ids.east) --node[pos=1, right, font=\labelFontSize]{$\bm H(\bm q)$} ++(\arrowLength,0);
\draw[output arrow] ([yshift=\blockHeight/6]ids.east) --node[pos=1, right, font=\labelFontSize]{$\bm C(\bm q, \dot{\bm q}) \dot{\bm q}$} ++(\arrowLength,0);
\draw[output arrow] (ids.east) --node[pos=1, right, font=\labelFontSize]{$\bm f(\dot{\bm q})$} ++(\arrowLength,0);
\draw[output arrow] ([yshift=-\blockHeight/6]ids.east) --node[pos=1, right, font=\labelFontSize]{$\bm g(\bm q)$} ++(\arrowLength,0);
\draw[output arrow] ([yshift=-\blockHeight/3]ids.east) --node[pos=1, right, font=\labelFontSize]{$\bm\tau$} ++(\arrowLength,0);

\node[square, below=0.5cm of ids, xshift=-1.25\blockHeight, inner sep=0pt] (ur10) {
    \includegraphics[height=0.95\blockHeight]{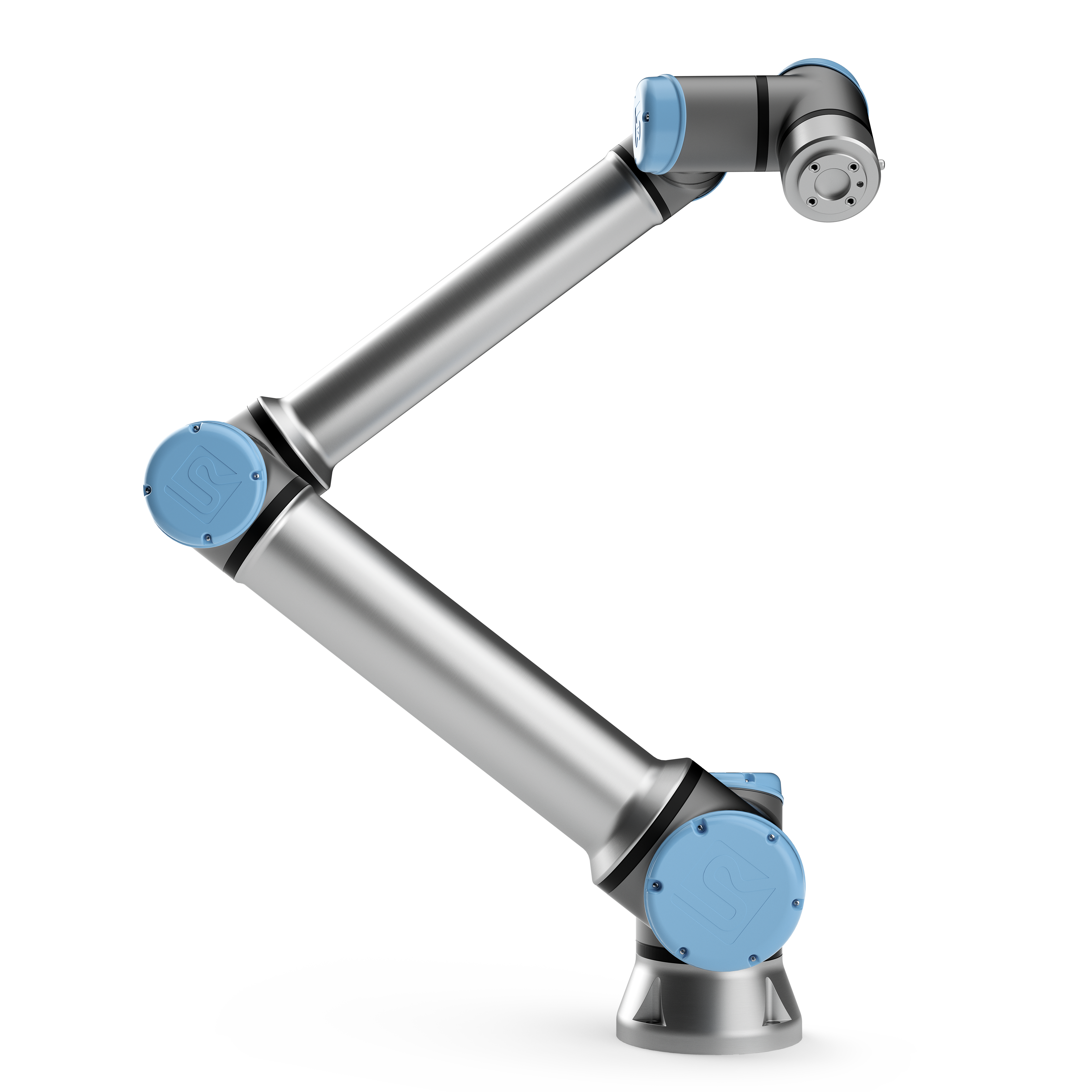}
};
\node[simple text, below=1pt of ur10] {UR10};
\node[square, below=0.5cm of ids, inner sep=0pt] (kdl) {
    \includegraphics[height=0.75\blockHeight]{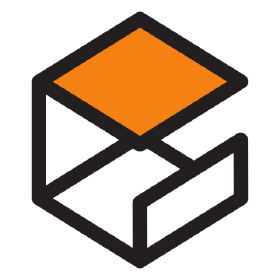}
};
\node[simple text, below=1pt of kdl] {KDL};
\node[square, below=0.5cm of ids, xshift=1.25\blockHeight, inner sep=0pt] (franka) {
    \includegraphics[height=0.95\blockHeight]{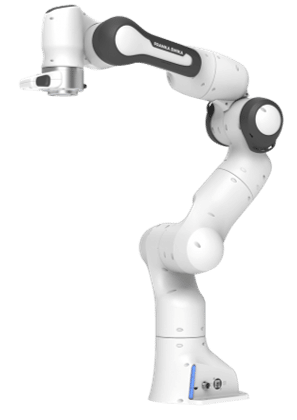}
};
\node[simple text, below=1pt of franka] {Franka};

\draw[arrow] ([xshift=-0.25\blockWidth]ids.south) -- ++(0,-0.25cm) -| (ur10.north);
\draw[arrow] (ids.south) -- (kdl.north);
\draw[arrow] ([xshift=+0.25\blockWidth]ids.south) -- ++(0,-0.25cm) -| (franka.north);

\end{tikzpicture}

\caption{The Inverse Dynamics Solver library and its concrete implementations}
\label{fig:inverse-dynamics}
\end{figure}

%% file: figures/ids-sequence-diagram.tex
\begin{figure*}
\centering
\includegraphics[width=\textwidth]{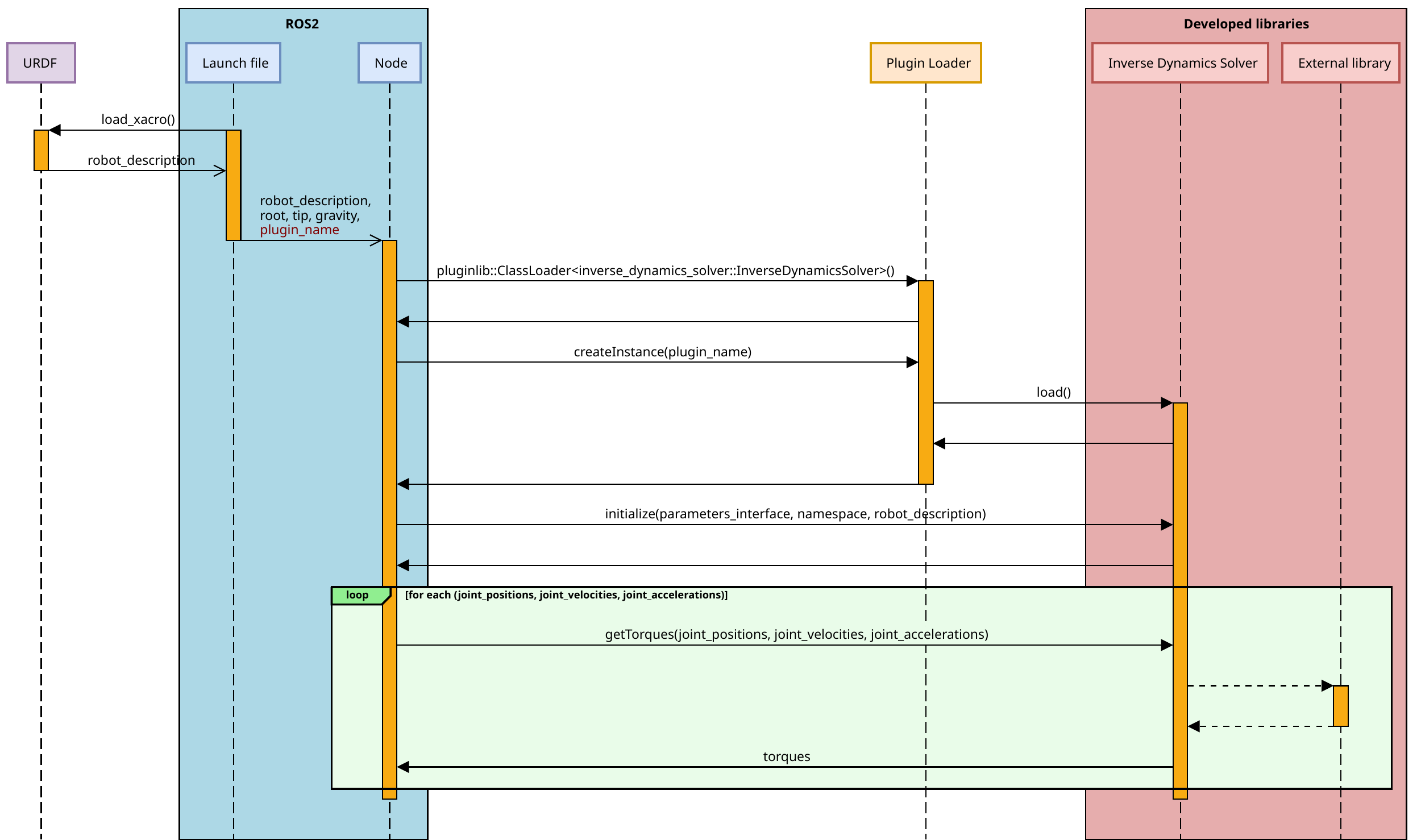}
\caption{Sequence diagram: the user can instantiate the IDS library with \texttt{pluginlib}, by specifying the parameter \texttt{plugin\_name} via launch file; concrete solvers call external libraries (\eg, robot-specific models \cite{gaz_dynamic_2019,petrone_dynamic_2024} or other software modules, such as KDL) to retrieve dynamic components.}
\label{fig:sequence-diagram}
\end{figure*}

%% file: figures/ids-class-diagram.tex
\begin{figure}
\centering
\begin{tikzpicture}
    
    \newcommand*{\eps}{0.02}  
    \newdimen\pad  
    \newdimen\imagewidth  
    \pad=\the\dimexpr \eps\columnwidth + \eps\columnwidth \relax
    \imagewidth=\dimexpr \columnwidth - \pad \relax

    \node[anchor=south west, inner sep=0] (image) at (0,0) {
        \includegraphics[width=\imagewidth]{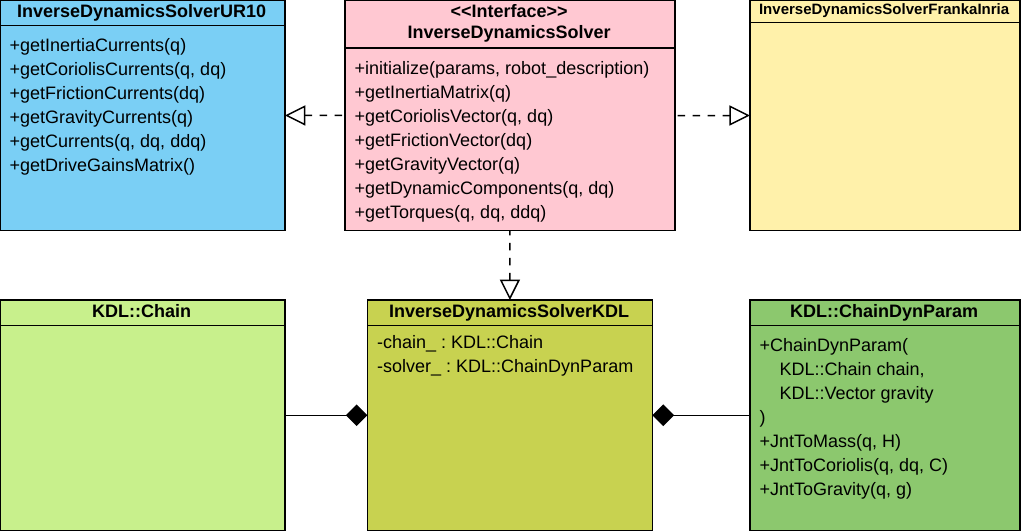}
    };
    \begin{scope}[x={(image.south east)}, y={(image.north west)}]
        \draw[dashed, rounded corners, draw=blue!50!black] 
            (-\eps, 1+\eps) --
            ++(1+2*\eps, 0) --
            ++(0, -0.5-\eps) --
            ++(-0.33-0.5*\eps, 0) --
            ++(0, -0.5-\eps) --
            ++(-0.34-\eps, 0) --
            ++(0, +0.5+\eps) --
            ++(-0.33-0.5*\eps, 0) --
            cycle;
        \node[fill=white, draw=blue!50!black, font=\tiny, inner sep=3pt, rounded corners] at (0.14, 0.5) {Proposed};
    \end{scope}
\end{tikzpicture}
\caption{Class diagram: methods inherited by the concrete implementations from the interface are omitted for brevity}
\label{fig:class-diagram}
\end{figure}

%% file: figures/kdl-ur10/kdl-ur10.tex
\begin{figure}
    \centering
    \newcommand*{\figsize}{0.47\columnwidth}

    \begin{center}
    \footnotesize{
    {\legblack} measured \hspace{0.5em}
    {\legred} computed
    }
    \end{center}
    \vspace*{-1em}

    \subfloat{
        \centering
        \includegraphics[width=\figsize]{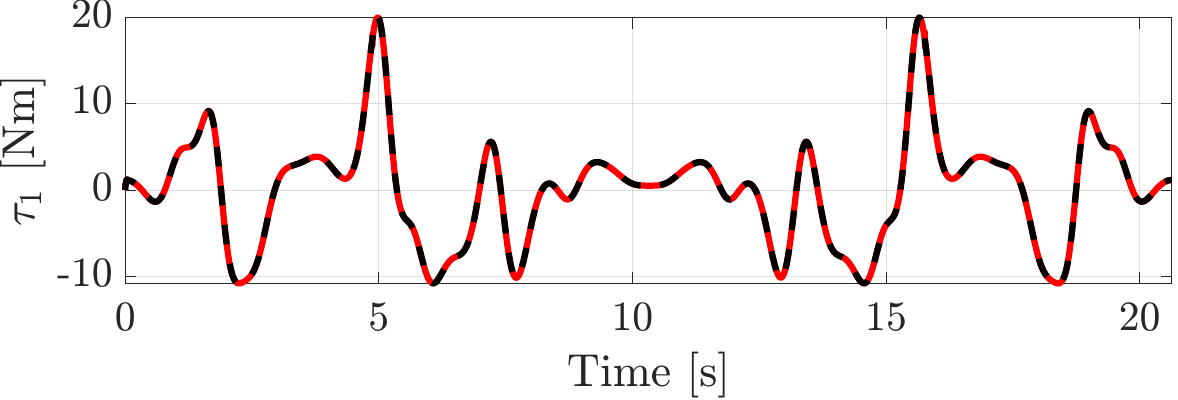}
    }\hfill
    \subfloat{
        \centering
        \includegraphics[width=\figsize]{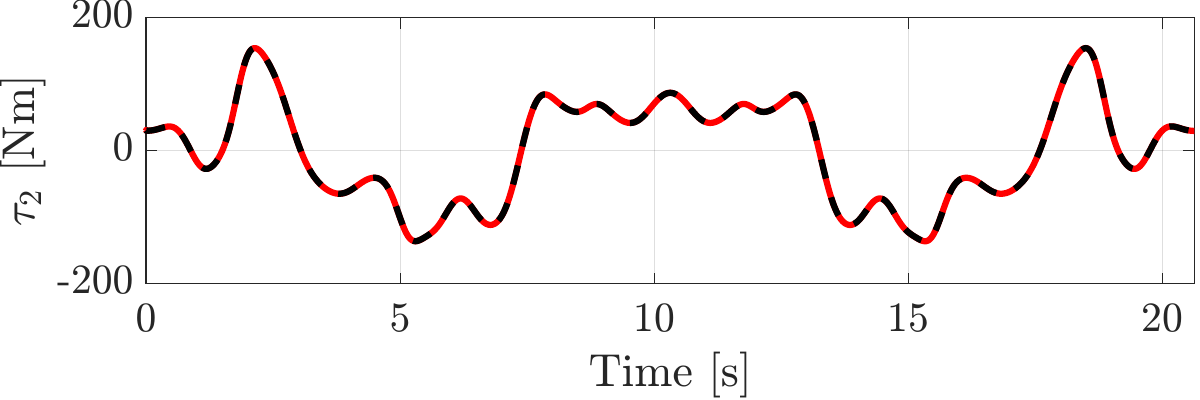}
    }\vspace*{-0.75em}\vfill
    \subfloat{
        \centering
        \includegraphics[width=\figsize]{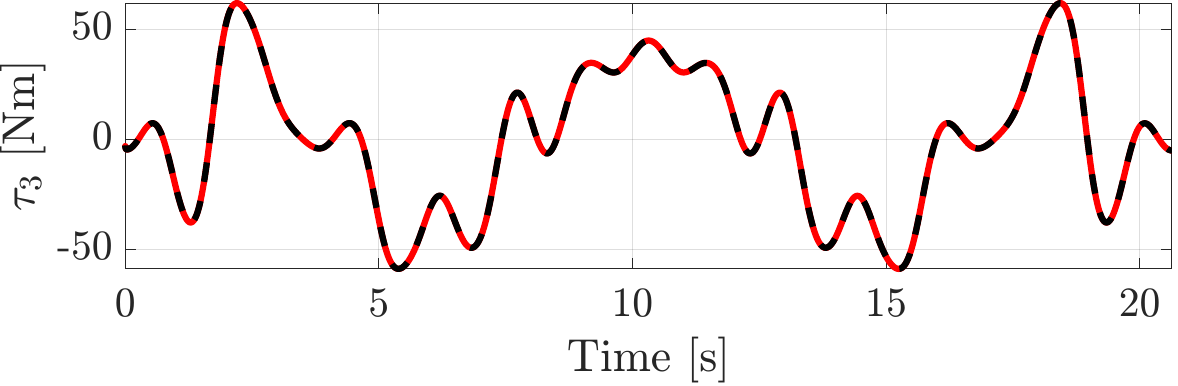}
    }\hfill
    \subfloat{
        \centering
        \includegraphics[width=\figsize]{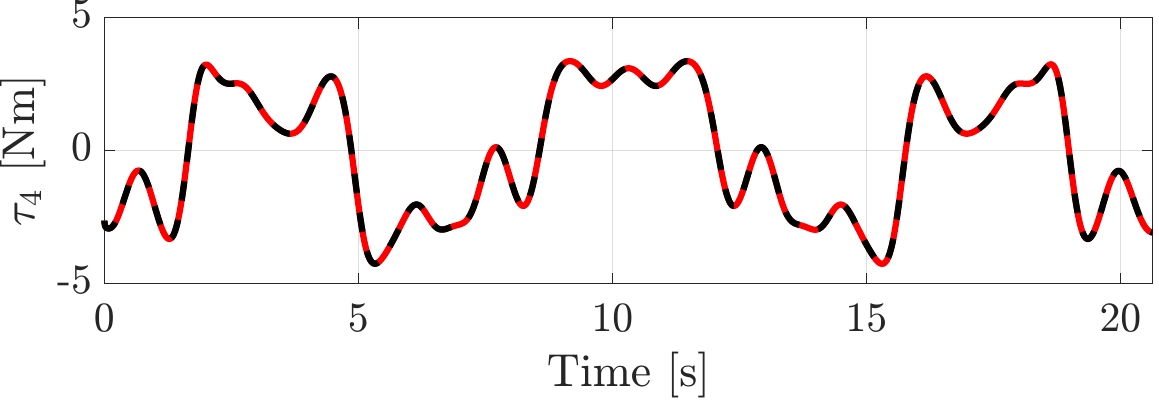}
    }\vspace*{-0.75em}\vfill
    \subfloat{
        \centering
        \includegraphics[width=\figsize]{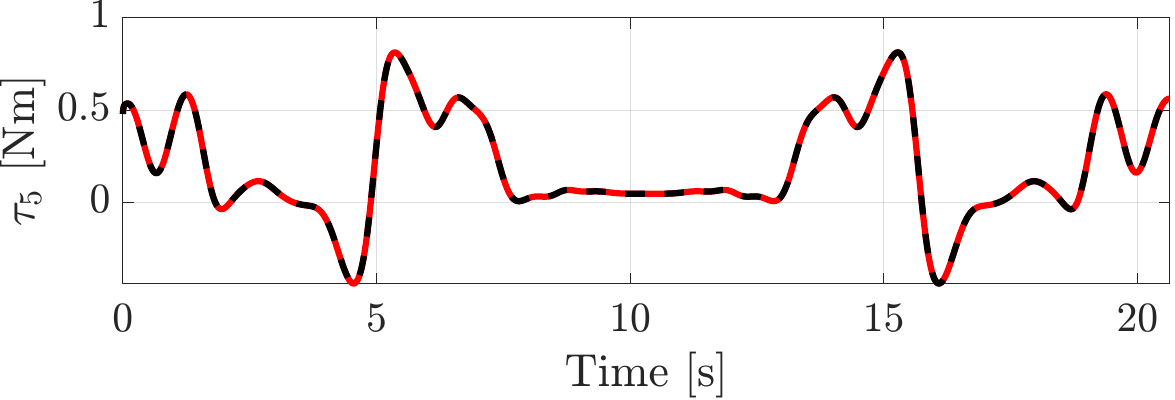}
    }\hfill
    \subfloat{
        \centering
        \includegraphics[width=\figsize]{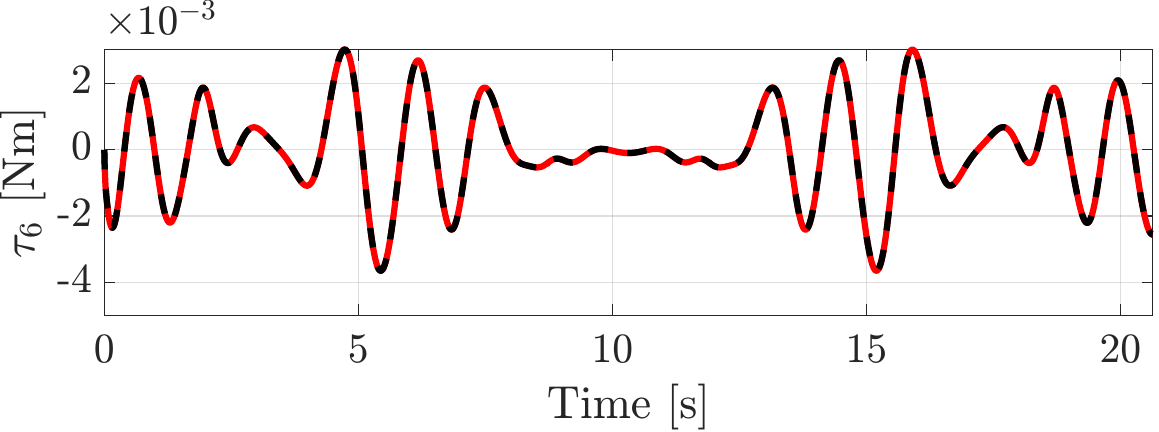}
    }
    \caption{Measured and computed torques on the simulated UR10 robot}
    \label{fig:kdl-ur10}
\end{figure}

%% file: figures/kdl-franka/kdl-franka.tex
\begin{figure}
    \centering
    \newcommand*{\figsize}{0.47\columnwidth}

    \begin{center}
    \footnotesize{
    {\legblack} measured \hspace{0.5em}
    {\legred} computed
    }
    \end{center}
    \vspace*{-1em}

    \subfloat{
        \centering
        \includegraphics[width=\figsize]{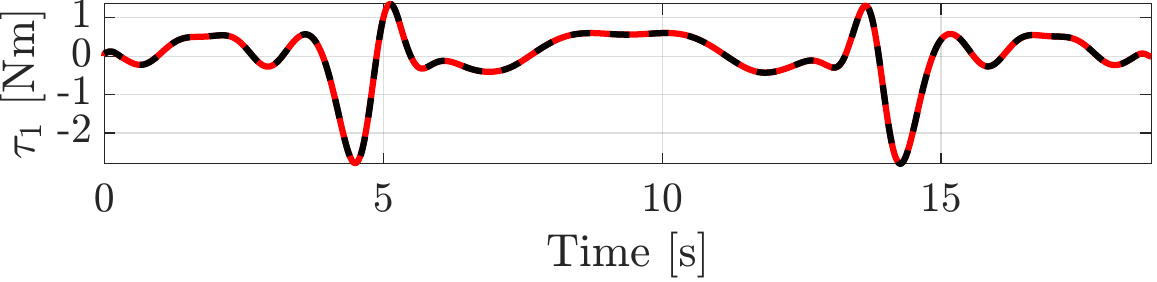}
    }\hfill
    \subfloat{
        \centering
        \includegraphics[width=\figsize]{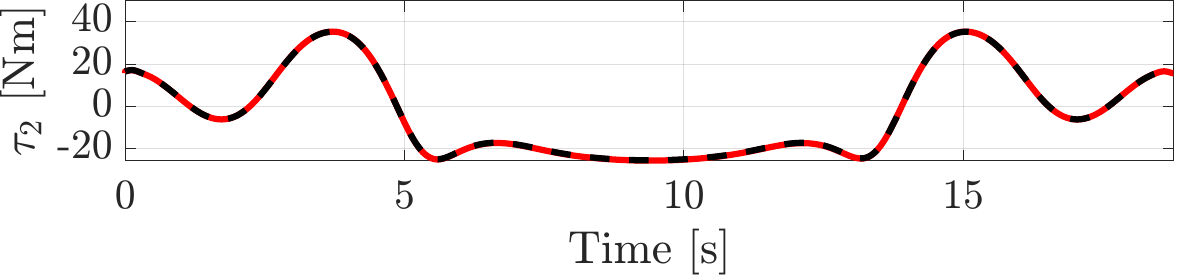}
    }\vspace*{-0.75em}
    \subfloat{
        \centering
        \includegraphics[width=\figsize]{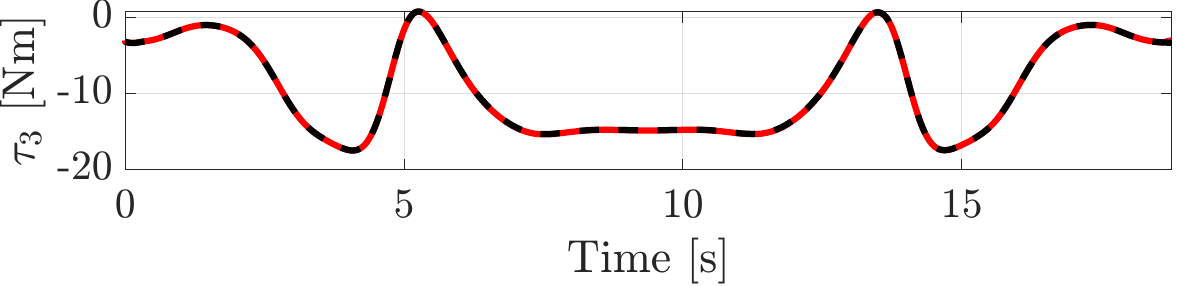}
    }\hfill
    \subfloat{
        \centering
        \includegraphics[width=\figsize]{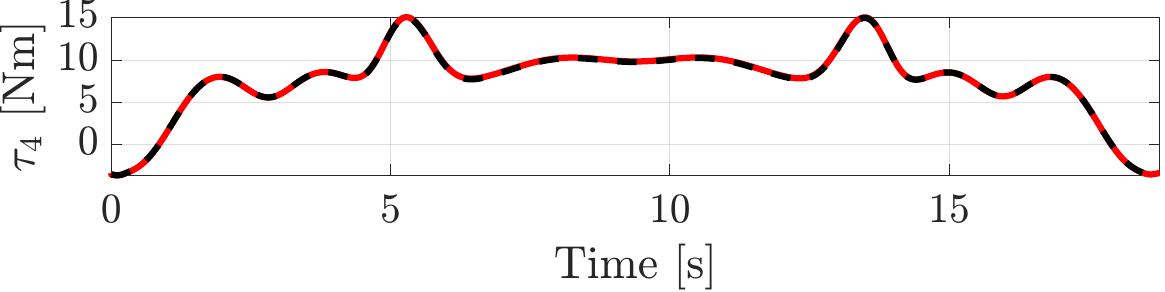}
    }\vspace*{-0.75em}\vfill
    \subfloat{
        \centering
        \includegraphics[width=\figsize]{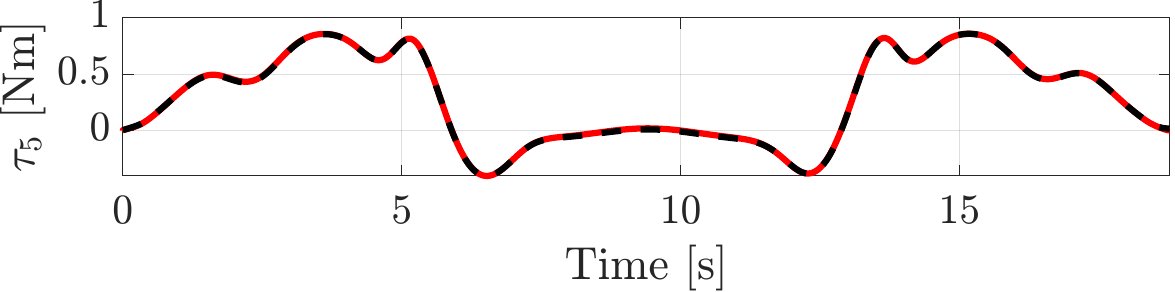}
    }\hfill
    \subfloat{
        \centering
        \includegraphics[width=\figsize]{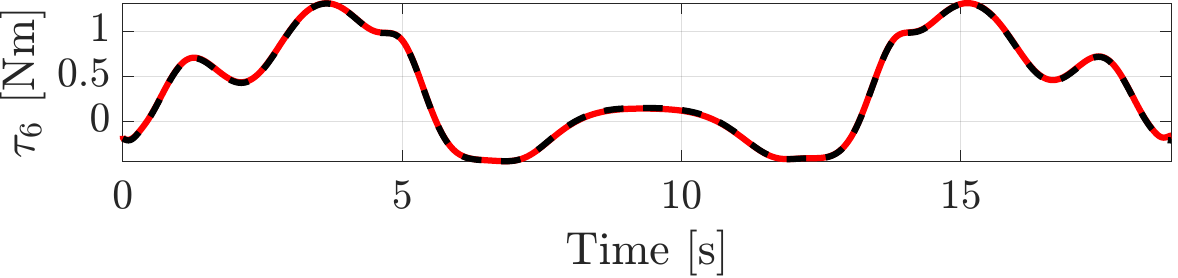}
    }\vspace*{-0.5em}
    \begin{center}
        \subfloat{
            \centering
            \includegraphics[width=\figsize]{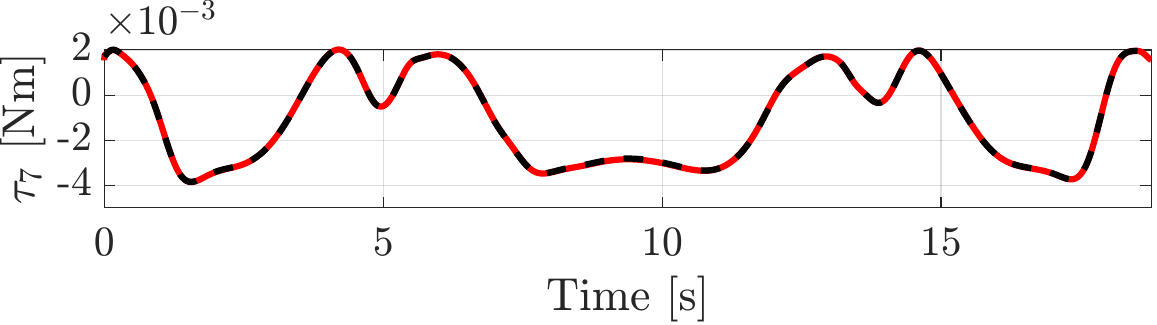}
        }
    \end{center}

    \caption{Measured and computed torques on the simulated Franka robot}
    \label{fig:kdl-franka}
\end{figure}

%% file: figures/ur10/ur10.tex
\begin{figure}
    \centering
    \newcommand*{\figsize}{0.47\columnwidth}

    \begin{center}
    \footnotesize{
    {\legblack} measured \hspace{0.5em}
    {\legred} computed
    }
    \end{center}
    \vspace*{-1em}

    \subfloat{
        \centering
        \includegraphics[width=\figsize]{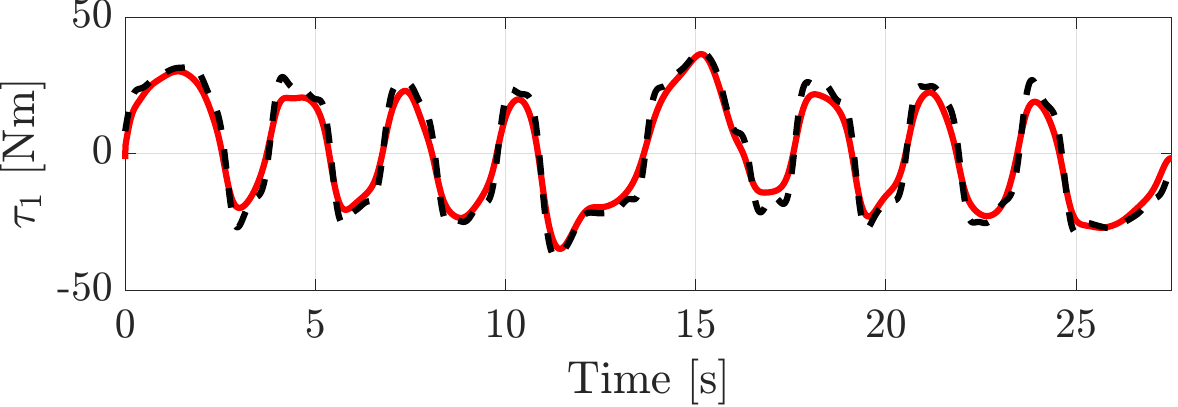}
    }\hfill
    \subfloat{
        \centering
        \includegraphics[width=\figsize]{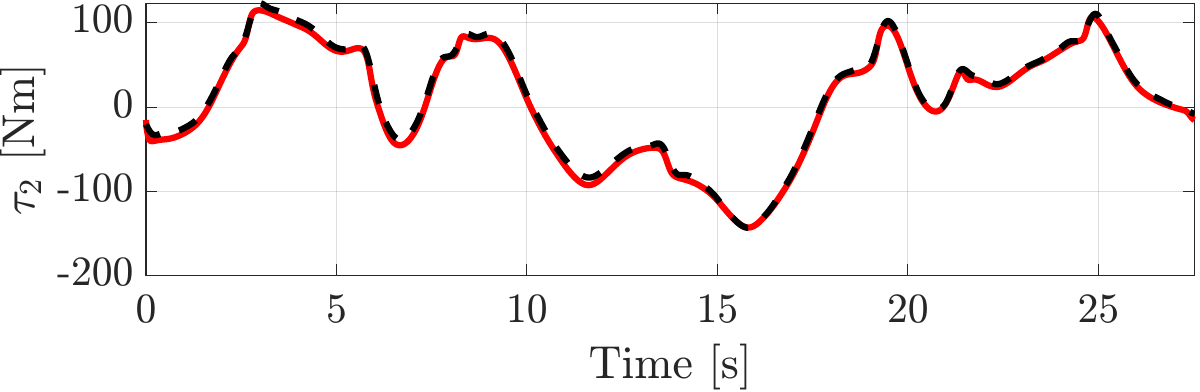}
    }\vspace*{-0.75em}\vfill
    \subfloat{
        \centering
        \includegraphics[width=\figsize]{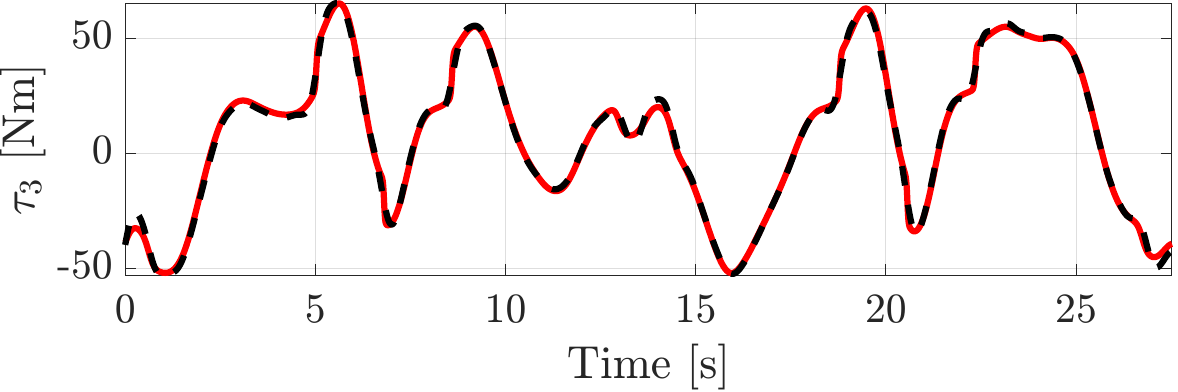}
    }\hfill
    \subfloat{
        \centering
        \includegraphics[width=\figsize]{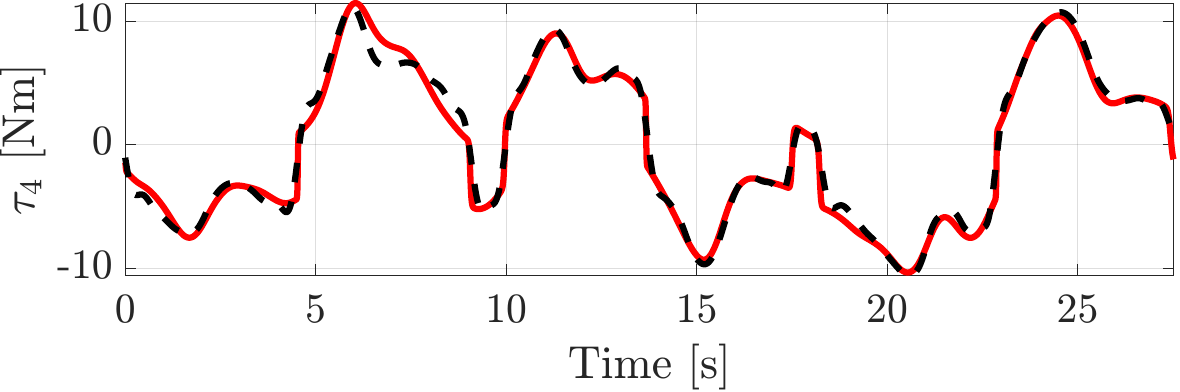}
    }\vspace*{-0.75em}\vfill
    \subfloat{
        \centering
        \includegraphics[width=\figsize]{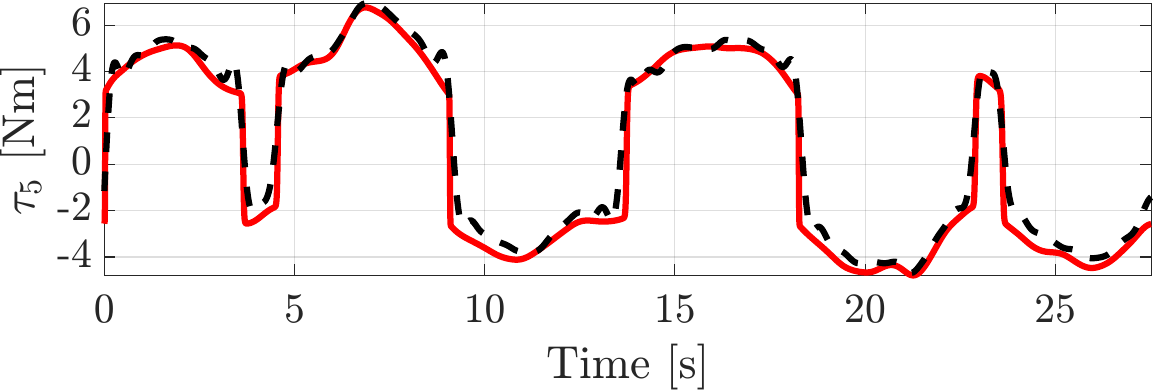}
    }\hfill
    \subfloat{
        \centering
        \includegraphics[width=\figsize]{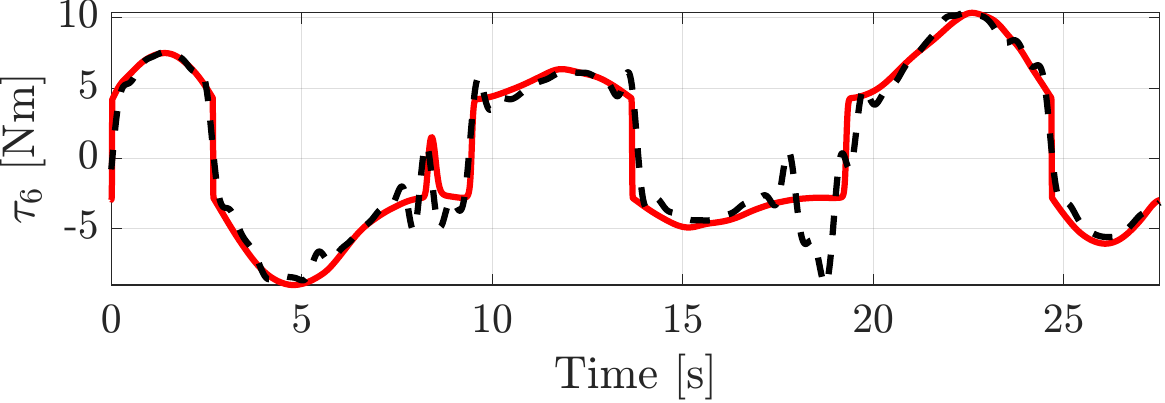}
    }

    \caption{Measured and computed torques on the real UR10 robot}
    \label{fig:ur10}
\end{figure}

%% file: figures/franka/franka.tex
\begin{figure}
    \centering
    \newcommand*{\figsize}{0.47\columnwidth}

    \begin{center}
    \footnotesize{
    {\legblack} measured \hspace{0.5em}
    {\legred} computed
    }
    \end{center}
    \vspace*{-1em}

    \subfloat{
        \centering
        \includegraphics[width=\figsize]{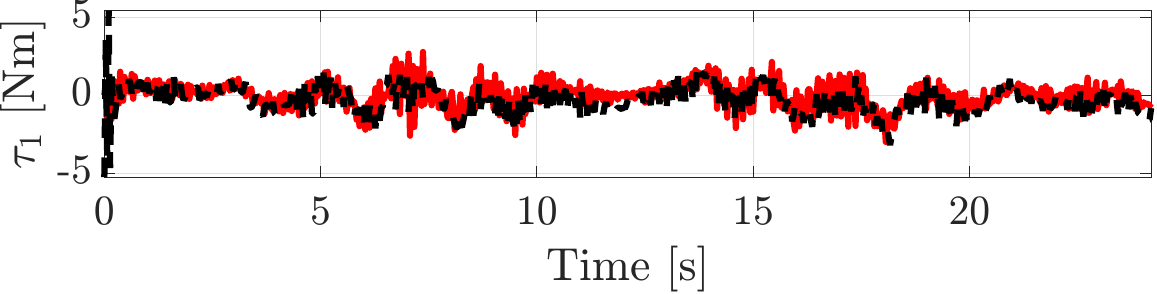}
    }\hfill
    \subfloat{
        \centering
        \includegraphics[width=\figsize]{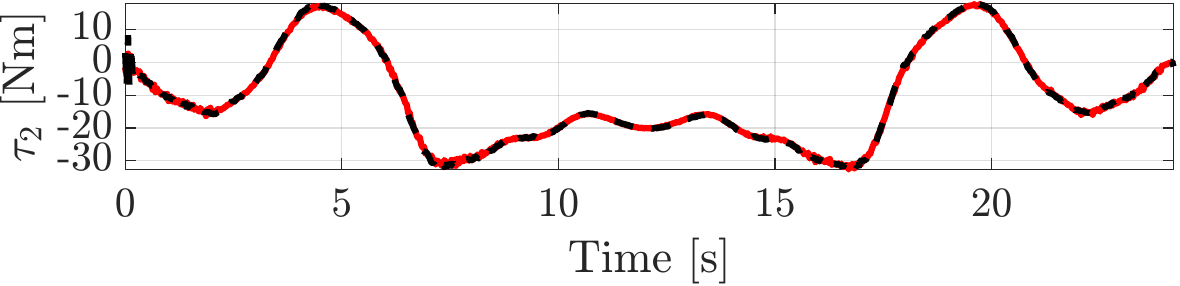}
    }\vspace*{-0.75em}\vfill
    \subfloat{
        \centering
        \includegraphics[width=\figsize]{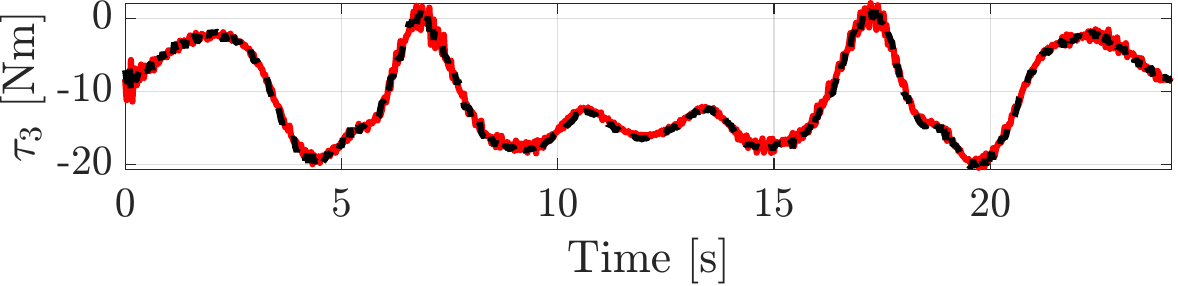}
    }\hfill
    \subfloat{
        \centering
        \includegraphics[width=\figsize]{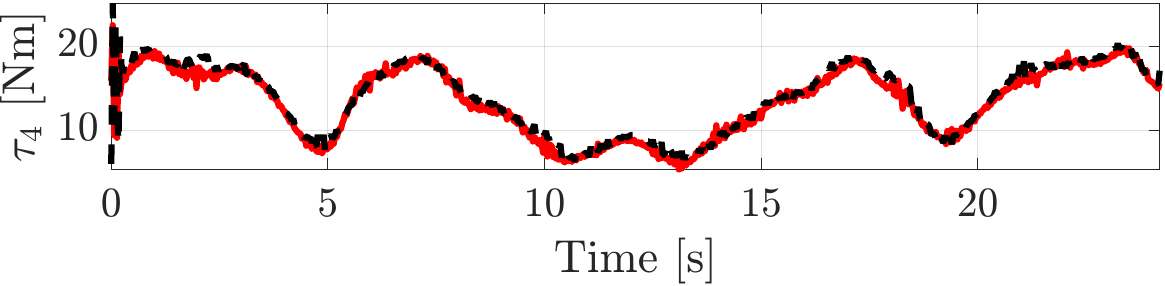}
    }\vspace*{-0.75em}\vfill
    \subfloat{
        \centering
        \includegraphics[width=\figsize]{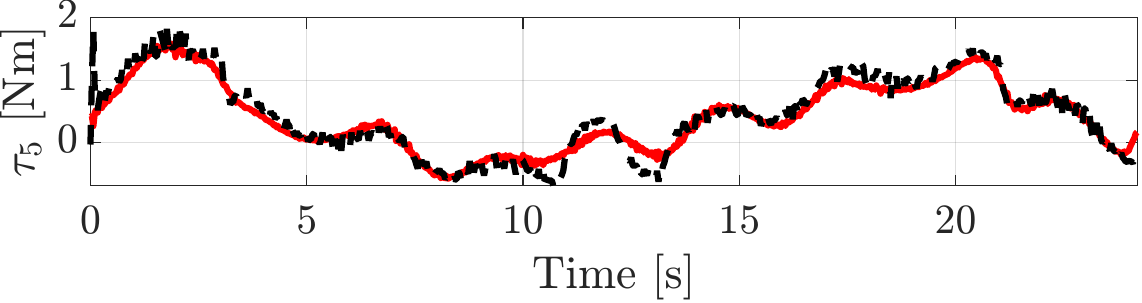}
    }\hfill
    \subfloat{
        \centering
        \includegraphics[width=\figsize]{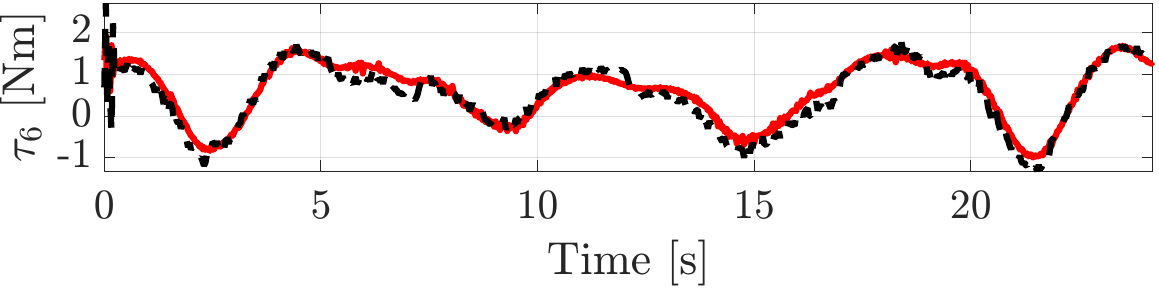}
    }\vspace*{-0.5em}\vfill
    \begin{center}
        \subfloat{
            \centering
            \includegraphics[width=\figsize]{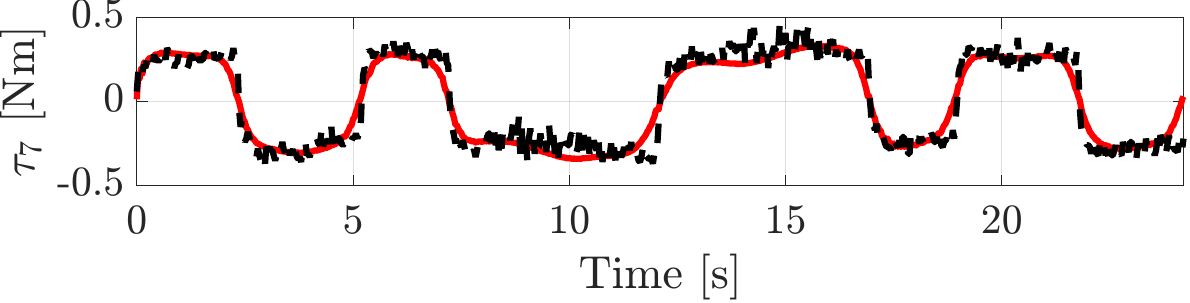}
        }
    \end{center}

    \caption{Measured and computed torques on the real Franka robot}
    \label{fig:franka}
\end{figure}

%% file: figures/robots/robots.tex
\begin{figure}
    \centering
    \newcommand*{\figsize}{0.47\columnwidth}

    \subfloat{
        \centering
        \includegraphics[width=\figsize]{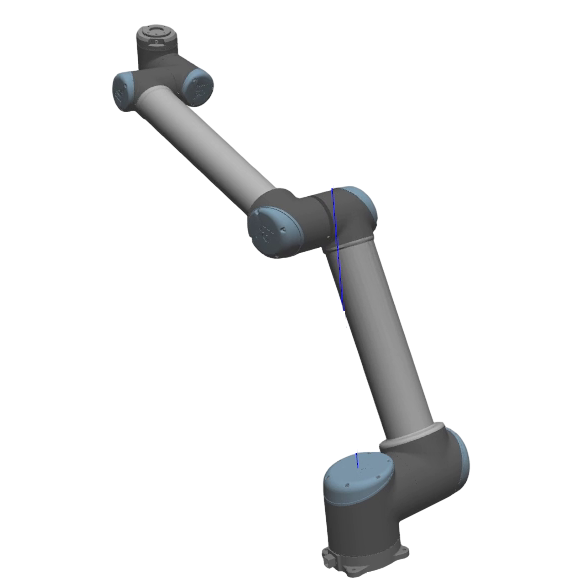}
    }\hfill
    \subfloat{
        \centering
        \includegraphics[width=\figsize]{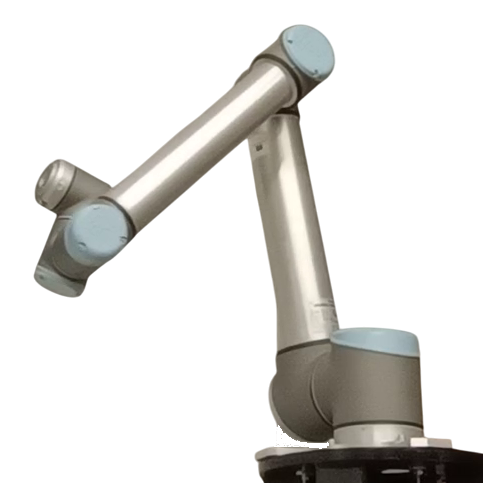}
    }\vfill
    \subfloat{
        \centering
        \includegraphics[width=\figsize]{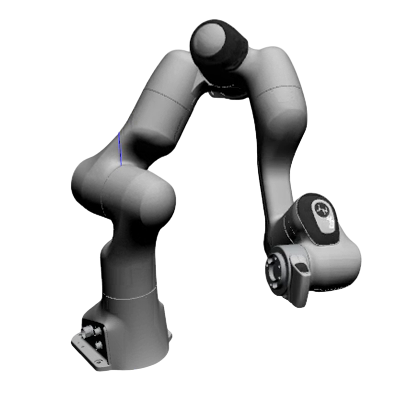}
    }\hfill
    \subfloat{
        \centering
        \includegraphics[width=\figsize]{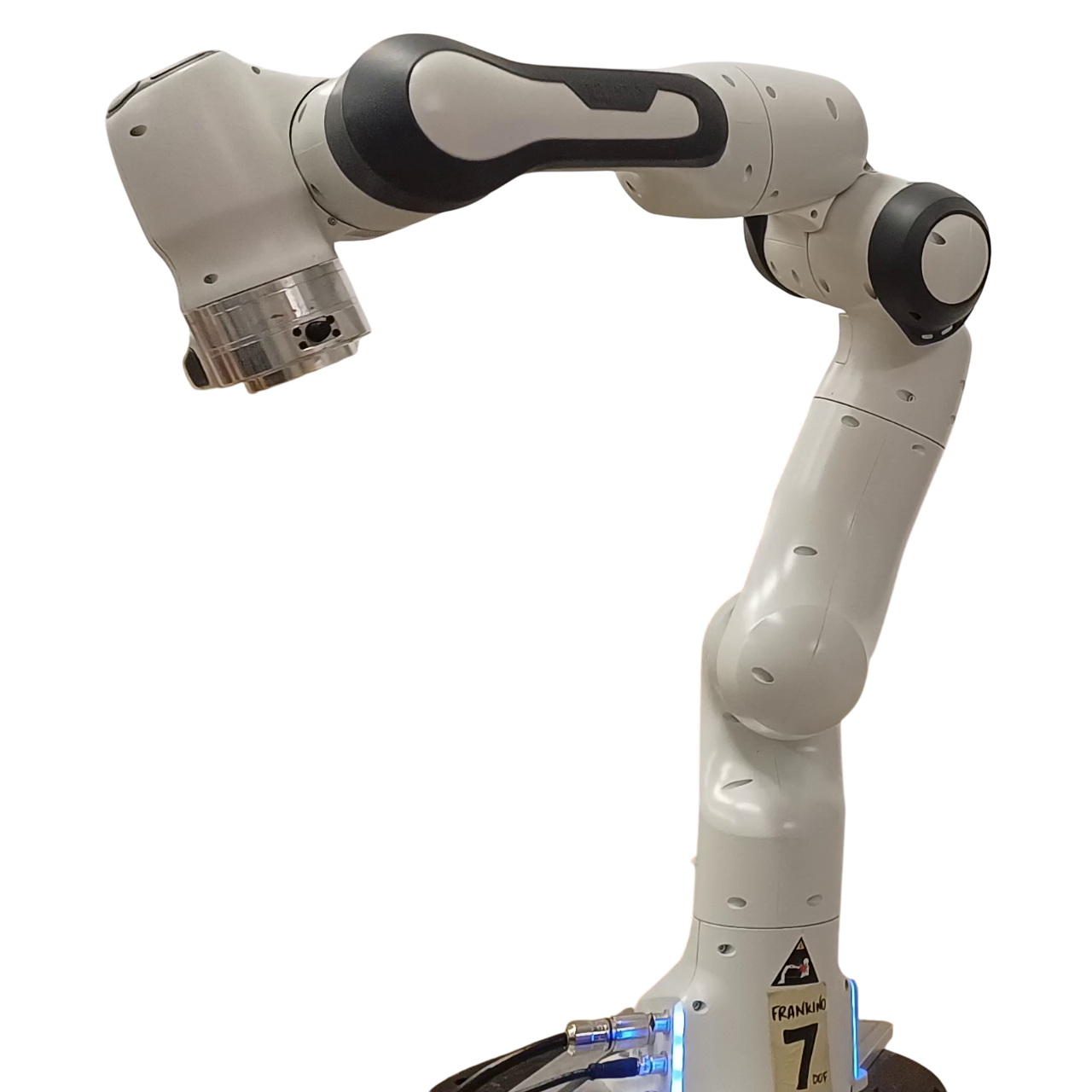}
    }

    \caption{Robots involved in the experiments: (top) UR10; (bottom) Franka; (left) simulated in Gazebo; (right) real laboratory setup}
    \label{fig:robots}
\end{figure}